\documentclass[letterpaper, 10 pt, conference]{ieeeconf}  

\IEEEoverridecommandlockouts                              

\usepackage{amsmath} 
\usepackage{amssymb}  
\usepackage{booktabs}
\usepackage{multirow}
\usepackage{graphicx}
\usepackage[table]{xcolor}
\usepackage{transparent}
\usepackage{comment}
\usepackage{threeparttable}
\usepackage{makecell}
\usepackage{pifont}
\usepackage{xfrac}

\title{\LARGE \bf
When to Adapt: Multi-Signal Domain Shift Detection for Efficient Training-Free Adaptation in Open-Vocabulary
Segmentation
}

\author{Michele Antonazzi$^{*\dagger}$, Alejandra C. Hernandez$^{*\ddagger}$, José Araujo$^\ddagger$, Olov Andersson$^\dagger$, Patric Jensfelt$^\dagger$ 
\thanks{$^*$Authors that contributed equally (randomly ordered).}
\thanks{$^\dagger$EECS and Digital Futures, KTH Royal Institute of Technology, Stockholm, Sweden. Corresponding author’s e-mail: \texttt{micant@kth.se}.}
\thanks{$^\ddagger$Ericsson Research, Ericsson AB, Stockholm, Sweden. Email: \texttt{name.surname@ericsson.com}.}
\thanks{This work was partially supported by the Wallenberg AI, Autonomous Systems and
Software Program (WASP) funded by the Knut and Alice Wallenberg Foundation.}%
}

\begin{document}

\maketitle

\thispagestyle{empty}
\pagestyle{empty}

\newcommand{\cellcolorfirst}[0]{\cellcolor[HTML]{1FF705}}

\newcommand{\cellcolorsecond}[0]{\cellcolor[HTML]{FFFFFF}}

\newcommand{\nadaptations}[0]{N$_\text{A}$}
\newcommand{\nadapters}[0]{N$_\text{U}$}

\newcommand{\aleja}[1]{{\leavevmode\color{black}#1}}
\newcommand{\michele}[1]{{\leavevmode\color{blue}#1}}
\newcommand{\ralaleja}[1]{{\leavevmode\color{magenta}#1}}

\newcommand{\final}[1]{{\leavevmode\color{green}#1}}

\newcommand{\best}[1]{\textbf{#1}}
\newcommand{\second}[1]{\underline{#1}}

\begin{abstract}
Robust and reliable perception is essential for autonomous robots operating in real-world environments, particularly in long-term missions where environmental conditions may change significantly over time. Although recent advances in Visual Foundation Models (VFMs) have improved open-vocabulary semantic segmentation, these models can still suffer from domain shift, which can significantly degrade performance if they are not adapted to the current environment. 
Training-free domain adaptation is a relevant paradigm for adaptation, consisting of adjusting the model online using lightweight adapters. Recent approaches apply this on a per-frame basis, which is impractical for deployments on resource-constrained robotic hardware.
To tackle this, we propose a multi-signal domain shift detection method for training-free continual test-time adaptation (TF-CTTA) in open-vocabulary segmentation. Our method leverages temporal coherence across consecutive frames by monitoring and combining complementary aspects of domain shift (visual change, adapter mismatch, and semantic drift) to trigger adaptation only when needed. We validate our approach on a benchmark including indoor and outdoor environments and using real robotic data. We demonstrate that our approach maintains segmentation accuracy while substantially reducing adaptations, making training-free adaptation practical and feasible for long-term, real-world robotic deployments.



  

\end{abstract}
\section{Introduction}





For autonomous robots operating in real-world environments, robust perception is essential, especially during long-term missions such as last-mile delivery, service robotics, or mobile assistance~\cite{perception_for_authonomy}. In these scenarios, a robot must navigate and understand scenes reliably across multiple and changing environments (e.g., alternating indoor and outdoor scenarios, new buildings, and varying illumination and weather). Such variability generates the well-known \emph{domain shift}, which causes performance degradation if the model is not adapted to the current operational environment~\cite{survey_domain_adaotation}.

Recent advances in Visual Foundation Models (VFMs), particularly for open-vocabulary semantic segmentation~\cite{sed}, have significantly improved robustness and generalization in out-of-distribution settings. Despite these zero-shot capabilities, VFMs can still suffer notable performance degradation under domain shift. Changes in scene type, appearance, lighting, or layout can lead to a noticeable decrease in the model's accuracy, making adaptation necessary during deployment~\cite{finetuning_vfm}. 

Although domain adaptation for semantic segmentation has been widely explored, a core challenge still remains:  how to adapt models online without introducing computational burden (e.g., operational cost or system complexity) while maintaining task performance. A potential solution is to rely on Continual Test-Time Adaptation (CTTA) approaches~\cite{hybrid_tta, cotta}, consisting of fine-tuning the model online after each inference step. However, this paradigm can be impractical in real-world robotic deployments because online fine-tuning via backpropagation introduces computational overhead, latency, and memory demands on resource-constrained devices. A promising alternative is training-free adaptation in which the model is adjusted at inference time without gradient-based updates. This paradigm has recently been introduced by SemLA~\cite{semla}, which proposes to dynamically fuse low-rank adapters into the model at test time, based on their proximity to the target domain. 

\begin{figure}[!t]
    \centering
    \includegraphics[width=\linewidth]{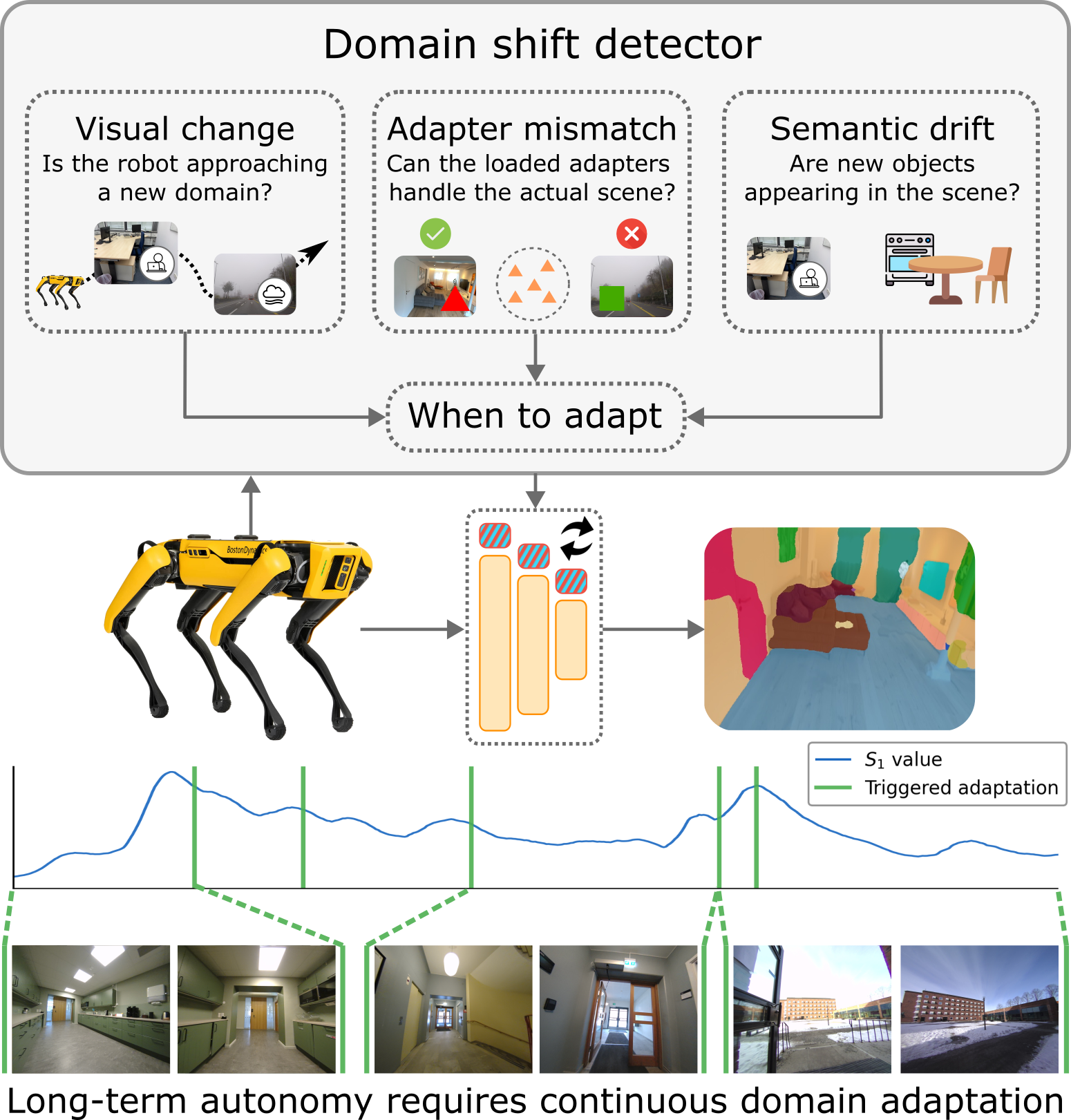}
    \caption{Multi-signal domain shift detection for training-free continual test-time adaptation. }
    \label{fig:intro}
    \vspace{-0.5cm}
\end{figure}
In this paper, we analyze SemLA for real-world robotic deployments, finding important practical limitations.
The method requires per-frame adapters retrieval and model updates, thus increasing energy consumption and introducing substantial runtime overhead, affecting real-time operation.
Furthermore, evaluations have mostly focused on general-purpose datasets, leaving evaluation on sequential robot data largely unexplored.

To avoid per-frame overhead, we propose a multi-signal domain-shift detector for identifying significant visual and semantic changes in image streams to trigger adaptation only when needed. While previous approaches relying on a single indicator (prediction entropy~\cite{emboditta} or feature statistics~\cite{simple_signal_domain_shift}) tend to saturate quickly and miss gradual changes, our approach detects shifts by monitoring three complementary aspects: (i) visual change, which captures abrupt scene transitions, (ii) adapter mismatch, which determines if the loaded adapters become inadequate for the current domain, and (iii) semantic drift, which identifies the appearance of novel semantic content. These signals are combined through a hierarchical fusion strategy that triggers adaptation only when multiple signals detect a domain shift.

Our framework, illustrated in Fig.~\ref{fig:intro}, integrates domain shift detection into a training-free continual test-time adaptation (TF-CTTA) pipeline for open-vocabulary segmentation used by mobile robots operating across indoor and outdoor environments. Rather than swapping adapters every frame like SemLA~\cite{semla}, our solution exploits temporal coherence across consecutive frames, triggering adaptation only when a domain shift is detected. This reduces adaptation events by two orders of magnitude while maintaining segmentation accuracy, making training-free adaptation practical for long-term robotic deployment.
In summary, our contributions are as follows:
\begin{itemize}
    \item We analyze the performance of SemLA~\cite{semla} on real-world robotic sequences, with a focus on indoor and alternating indoor-outdoor scenarios, identifying practical limitations of per-frame adaptation.
    \item We introduce a domain shift detector that combines three complementary signals to guide the adaptation, reducing the overhead caused by per-frame adaptation while maintaining semantic segmentation quality.
    \item We propose a continual test-time adaptation benchmark tailored to robotics, capturing alternating and temporally correlated domain shifts\footnote{The code and the benchmark will be released upon publication.}. 
    Using this, we provide a comprehensive evaluation comparing our method for TF-CTTA with relevant baselines.
    \item We demonstrate our approach on real-world sequences acquired with a robotic sensor across indoor and outdoor environments and benchmark inference on NVIDIA Jetson hardware.
\end{itemize}

\section{Related Work}




\textbf{Continual test-time adaptation (CTTA)} is a challenging task that focuses on enabling models to adapt in real-time to dynamic, changing environments, without access to the source data. 
This task was initially introduced in the seminal work of~\cite{cotta}, where a neural network is fine-tuned online with the supervision of a teacher, which is updated as an exponential moving average of the student's weights.
EcoTTA~\cite{ecotta} improves the teacher-student framework using lightweight meta networks that are attached to the original frozen model, while ViDA~\cite{vida} leverages visual domain adapters added to a source-pretrained model.
Moving toward more realistic deployment scenarios, \cite{multimodal_ctta} studies multi-modality test-time adaptation for semantic segmentation, leveraging multiple sensing modalities to improve robustness under domain shifts and highlighting the relevance of methods tailored for robotic perception. 
Despite attractive, the aforementioned solutions present important limitations for real-world deployment, in terms of both inference speed and memory consumption, as they require backpropagation and weight sharing for EMA accumulations between two large architectures.

\textbf{Training-free continual test-time adaptation (TF-CTTA)} is a different paradigm that avoids expensive parameter updates by combining pretrained models relevant to the current domain, avoiding any additional training.
\cite{uniform_adapter_merging} proposes a model merging technique, including parameter and buffer merging for multi-target domain adaptation tasks, applicable to any single-target domain adaptation model. Most closely related to our work, \cite{semla}~introduces SemLA, a training-free test-time framework for open-vocabulary segmentation that leverages a library of LoRA-based adapters indexed with Contrastive Language-Image Pretraining (CLIP) embeddings~\cite{clip}. The CLIP embeddings guide adapter selection, and subsequently, the most relevant adapters are merged based on their proximity to the target domain in the embedding space. Despite being intriguing, updating the model every frame strongly compromises real-time inference, remaining not practical for real-world robotic deployments.

\textbf{Domain shift detection} represents a valuable technique for reducing the model updates by identifying substantial changes in the robot's perception stream that can cause performance degradation. In continual test-time adaptation, a natural solution is to leverage the discrepancy between the teacher-student predictions. The works of~\cite{hybrid_tta} and~\cite{real_time_ctta} compare the consistency loss between teacher and student accumulated with EMA with a variable threshold. 
\cite{ccotta} models domain shift directions via Concept Activation Vectors (CAVs)  and incorporates domain shift and class shift controllers to prevent category drifting.

Other approaches detect domain shifts by performing uncertainty estimation using a model trained on the source domain. 
In ViDA~\cite{vida}, each sample is processed multiple times using dropout in the linear layers, and uncertainty is calculated as the variance across the outputs. 
Similarly,~\cite{ctta_dynamic_selection} computes a threshold using an EMA by aggregating the maximum confidence values for each pixel calculated using multiple versions of the same sample obtained with data augmentation. The approaches of~\cite{onda} and~\cite{dual_ctta} estimate uncertainty using the average class-wise confidence of a source model accumulated in a sliding window with a threshold to identify domain changes. 
Teacher-student and source-model uncertainty methods are impractical in our scenario, as deploying an additional segmentation model or running multiple forward passes for uncertainty estimation significantly compromises computational and memory efficiency.

Domain shifts can also be detected by monitoring features or prediction statistics directly, with no secondary model. \cite{simple_signal_domain_shift}~proposes a domain shift signal based on the cosine distance between mean feature representations of consecutive batches from the source model, triggering when the signal exceeds a multiple of its moving average. More recently, \cite{emboditta}~targets resource-constrained embodied devices, monitoring prediction entropy via EMA and adapting only when entropy deviates past a threshold.
Despite being more efficient, these methods are not robust enough for long-term sequences of robotic perceptions. Relying on a single indicator, they tend to saturate quickly as they cannot differentiate between types of domain shift. Our method, instead, combines multiple complementary signals for capturing different types of shifts and, relying on CLIP features, is specifically designed for the training-free domain adaptation framework of~\cite{semla}.

\begin{figure*}
    \centering
    \includegraphics[width=1\linewidth]{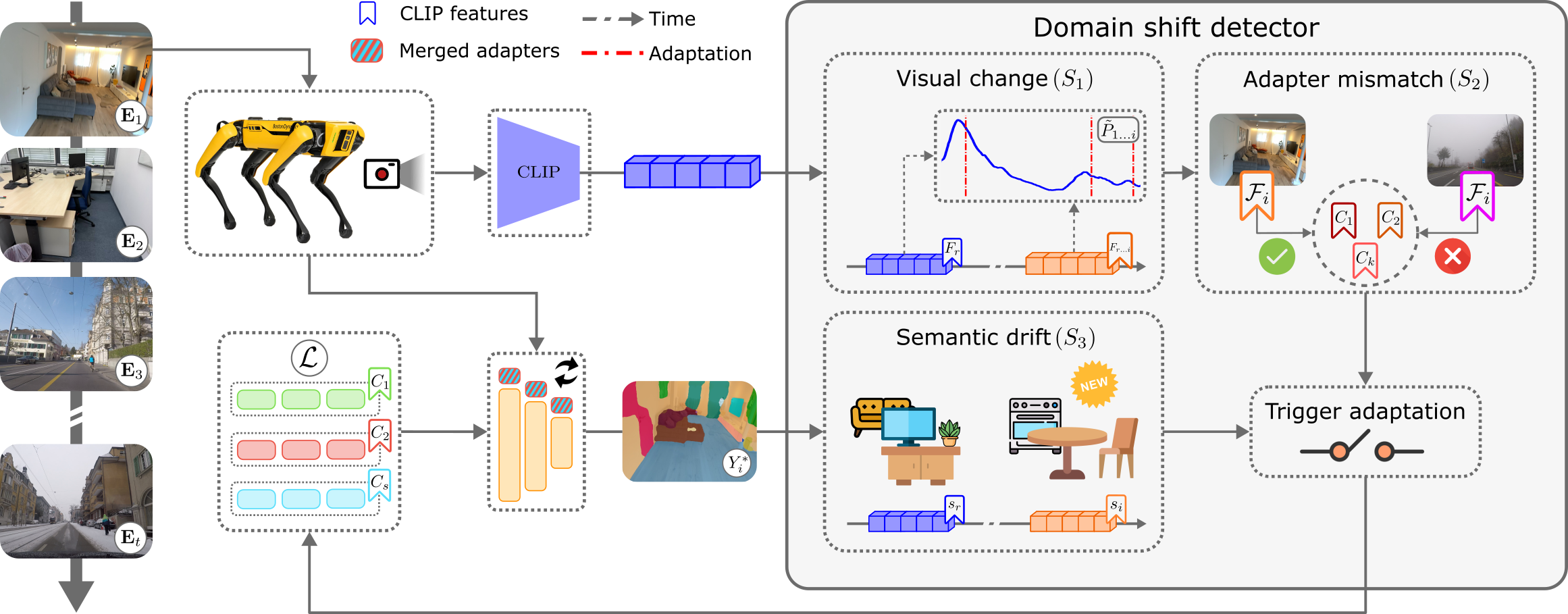}
    \caption{General overview of the framework for training-free continual test-time adaptation, incorporating SemLA~\cite{semla} and our domain shift detector.}
    \label{fig:method}
\vspace{-0.5cm}
\end{figure*}

\textbf{Positioning our work.} Our work brings training-free continual test-time adaptation (TF-CTTA) to real-world robotic deployments. Considering the sequential image streams perceived by robots in the wild, we propose a multi-signal domain shift detector that leverages temporal coherence to trigger adaptation only when a domain shift is detected, improving efficiency while maintaining segmentation accuracy.


\section{Preliminaries}

\subsection{Problem formulation}\label{sec:method:problem_formulation}

In this paper, we consider a robot performing open-vocabulary semantic segmentation from images. More formally, this task can be defined as $Y^* = f_{\theta_p}(I)$, where a neural network $f$ correlates an image $I$ to a dense mask $Y^*$, where each pixel is assigned to an object category. The network's parameters $\theta_p$ are pre-trained using a labeled \emph{source} dataset $\mathcal{P} =(\mathbf{I}^p, \mathbf{Y}^p, \mathbf{O}^p)$ containing a list of images ($\mathbf{I}^p$) and their corresponding ground truth masks ($\mathbf{Y}^p$), where object categories come from $\mathbf{O}^p$. 
Inside the same mission, a robot needs to operate in a sequence of \emph{target} environments $\mathcal{T} = [\mathbf{E}_t\:|\: t= 1,\ldots, T]$ that can be both indoor or outdoor. Inside each environment $\mathbf{E}_t= \{\mathbf{I}_t, \mathbf{O}_t\}$, the robot is tasked to segment a sequence of RGB images $\mathbf{I}_t$ using a specific set of object categories $\mathbf{O}_t$. Note that we are working in open-vocabulary settings, so the classes used for initializing the models and those required at deployment time may differ.


Training-free continual test-time adaptation (TF-CTTA) aims to adapt the neural network $f_{\theta_p}$, initially pre-trained on the source dataset $\mathcal{P}$, to a sequence of target datasets $\mathcal{T}$, thus obtaining a new model $f_{\theta_t}$ with improved performance in the domains seen by the robot during deployment. 
This process is constrained by three main challenges: (i) ground-truth labels of the target data are unavailable, (ii) the model must evolve online using the stream of images in $\mathcal{T}$, and (iii) the adaptation must be training-free, meaning that fine-tuning using backpropagation is not allowed. 

Fig.~\ref{fig:method} presents a general overview of the approach we propose for enabling training-free continual test-time adaptation in mobile robots. The framework builds on the training-free approach of SemLA~\cite{semla}, summarized in Section~\ref{sec:method:semla}.
The core of our contribution, detailed in Section~\ref{sec:method:shift_detection}, is a domain
shift detection method based on CLIP features to identify when adaptation is needed. By preventing unnecessary adaptation, our method reduces the number of adaptations while preserving segmentation performance, making the approach more efficient and practical for long-term, real-world deployments.

\subsection{Training-free adaptation via adapters merging}\label{sec:method:semla}

Our domain shift detection method adapts the training-free adaptation proposed by SemLA~\cite{semla} to online robotic deployment. SemLA encodes domain-specific knowledge in adapters and dynamically merges the most relevant ones based on proximity in CLIP's~\cite{clip} embedding space.

Let us consider a pre-trained model $f_{\theta_p}$ and a set $\mathcal{S} = \{\mathbf{S}_s\: |\: \mathbf{S}_s = \{\mathbf{I}_s, \mathbf{Y}_s, \mathbf{O}_s\} \}_{s=1}^S$ containing $S$ \emph{source} datasets.
SemLA is a library of adapters formally described as
$$\mathcal{L}=\{(C_s, \theta_s)\:|\: s= 1,\ldots,S\},$$
where $C_s$ is the descriptor for the dataset $\mathbf{S}_s$ (i.e., the average of the CLIP features from the images $\mathbf{I}_s$), while $\theta_s$ denotes the weights of a low-rank adapter~\cite{lora} trained on $\mathbf{S}_s$.
%
At deployment time, for each environment $\mathbf{E}_t \in \mathcal{T}$ and each image $I_i \in \mathbf{I}_t$, SemLA first computes the CLIP embedding $F_{i}$ for the test image. Then, this embedding is used to retrieve the top-$K$ most relevant adapters from the library $\mathcal{L}$, namely those adapters whose centroids are closest to the image's embedding in terms of Euclidean distance.
The distances between the image and the adapters descriptors are then normalized using a softmax function and used to scale the adapters weights, that are then merged and attached to the pre-trained model before inference on $I_i$.


\section{Multi-Signal Domain Shift Detection}\label{sec:method:multisignal}
The main limitation of applying SemLA on mobile robots is its per-frame adaptation. We observed that this requirement significantly compromises computational efficiency, as the model needs to be dynamically updated every iteration.
Since robotic perceptions operating in an environment are highly correlated, we overcome this limitation by proposing a multi-signal domain shift detection to trigger adaptation only when needed. Specifically, our method monitors three signals capturing complementary aspects of domain shift: the visual change ($S_1$), whether the loaded adapters still match the current domain ($S_2$), and the drift in the semantic objects appearing in the stream ($S_3$). The signals are fused hierarchically to trigger adaptation only when substantial evidence is detected across them, thus avoiding costly per-frame adaptation and making training-free adaptation practically deployable in robotic hardware.


\subsection{Visual change signal ($S_1$)}\label{sec:method:shift_detection}
%


The visual change signal $S_1$ captures global appearance changes by monitoring how CLIP embeddings of incoming frames diverge from the current domain representation. It answers the question: has the environment changed visually?

Consider the ordered stream of images contained in all the target environments of $\mathcal{T}$, defined as $\mathbf{I}_\mathcal{T}$, with length $|\mathbf{I}_\mathcal{T}| = \sum_{t=1}^T|I_t|$ equal to the sum of the number of images of all environments $\mathbf{E}_t \in \mathcal{T}$.

At each frame $i$, we use CLIP to compute a feature embedding $F_{i}$ for the image $I_i$ and measure its proximity to a feature memory $\mathcal{F}_{i-1}$, which encodes the visual appearance of the image stream up to step $i-1$. The memory is updated using an exponential moving average (EMA) of the incoming features until the latest step $i-1$, formally $\mathcal{F}_{i-1} = (1-\alpha)\mathcal{F}_{i-2} + \alpha F_{i-1}$, where $\alpha=10^{-2}$. The proximity score between the memory $\mathcal{F}_{i-1}$ and the current feature $F_i$ is calculated as $P_i=1- \text{cos}(\mathcal{F}_{i-1}, F_i)$, where $\text{cos}(\cdot , \cdot)$ is the cosine similarity.

The signal works by tracking the temporal evolution of $P_i$ over time: 
an \emph{increasing} trend suggests a domain change, while a \emph{decreasing} trend indicates a stable domain.
Let $P_{1\ldots i}$ denote the proximity scores from the first to the current frame $i$. 
At first, we apply a low-pass filter to obtain a smoothed signal, formally $\tilde{P}_{1\ldots i} = G_{(\sigma, w)} \ast P_{1\ldots i},$ where $(\ast)$ is the convolution operation and $G_{(\sigma, w)}$ is a Gaussian kernel with standard deviation $\sigma = 6$ and amplitude $w$. Then, at the current frame $i$, we compute the derivative of $\tilde{P}_{w\ldots i}$ considering only the last $i-w+1$ frames. It is calculated as 
$D_i = S_{(w)} \ast \tilde{P}_{w\ldots i},$
where $S_{(w)}$ is a Sobel kernel applied in window $w$. 
Since single derivatives are noisy, the trend is encoded in a memory variable $\mathcal{D}$, which accumulates derivatives using EMA (with $\alpha=10^{-2}$). 

The trend of the signal is monitored using the derivative memory $\mathcal{D}_{i-1}$ until the last frame $i-1$. When the domain is changing, two conditions must be satisfied. First, the memory must be positive ($\mathcal{D}_{i-1} >0$), indicating an increasing trend. Secondly, the current derivative must substantially exceed the memory, i.e., $D_i>\rho\mathcal{D}_{i-1}$, where $\rho$ is a sensitivity hyperparameter (fixed to $\rho = 10$).
Instead of triggering the signal immediately, we wait until the derivative memory becomes negative ($\mathcal{D}_{i-1}<0$), indicating that the signal trend is decreasing. The visual change $S_1\in\{0,1\}$ is then a binary signal.  

\subsection{Adapter mismatch signal ($S_2$)}\label{sec:method:adapter_drift}
A visual change does not necessarily mean that the active adapters are inadequate: a scene may change in appearance while remaining within the adapters' competence. Conversely, gradual shifts in layout or viewpoint can cause an adapter mismatch without any detectable visual change. $S_2$ asks: do the active adapters still match the current domain?

To tackle this, we introduce an adapter mismatch signal that monitors how far the current domain has moved from the active adapters' competence regions. This score is formally computed as $\delta_i = \min_{k=1}^{K} 1-\text{cos}(\bar{F}_i, C_k)$, i.e., the lowest inverse cosine distance score between the representation of the current domain $\bar{F}$ (the average of the CLIP features from the last $w$ frames) and the top-$K$ loaded adapters. The rationale is that if $\delta_i$ is low, then there is at least one adapter well aligned with the current domain, while it increases only when all the adapters become unsuitable. We record the $\delta_{\text{base}} = \delta_i$ at the first frame after each adaptation and we define 
$
    S_2(i) = \tfrac{\delta_i}{\delta_{\mathrm{base}}} > \tau_2.
$
When $S_2(i) > \tau_2$, the adapters no longer cover the current domain as well as they did right after being loaded. Normalizing by $\delta_{\mathrm{base}}$  makes $\tau_2$ independent of absolute embedding distances, so the same threshold works across environments.
\subsection{Semantic drift dignal ($S_3$)}\label{sec:method:semantic_drift}
$S_1$ and $S_2$ operate on CLIP image embeddings and ignore the model's predictions. Two visually distinct scenes may share the same semantic composition, while two visually similar ones may differ in object content. $S_3$ answers a different question: has the semantic content changed relative to what the active adapters were trained for?

For each frame $i$, we compute a semantic descriptor 
$\mathbf{s}_i = \text{norm}_{2}(\mathbf{c}_i^\top \mathbf{E})$,
where $\text{norm}_2(\cdot)$ is $L_2$-normalization, $\mathbf{c}_i \in \mathbb{R}^{|\mathbf{O}|}$ is the vector of normalized pixel frequencies for each object class in $\mathbf{O}$ for the prediction $i$, while $\mathbf{E} \in \mathbb{R}^{|\mathbf{O}| \times d}$ is the matrix of pre-computed $L_2$-normalized CLIP text embeddings for the object categories $\mathbf{O}$ ($d$ is the embedding dimension). 
We maintain a memory bank $\mathcal{M} = \{(\mathbf{s}^{m},\, \mathcal{O}^{m})\}_{m=1}^{M}$ storing the semantic context of $M$ adaptation events (we fix $\text{M}=50$). Each entry consists of: (i) a semantic descriptor $\mathbf{s}^{m}$ computed as the mean of the $W_s$ semantic scores after adaptation, and (ii) a class set $\mathcal{O}^{m}$, the union of the $k_c$ most frequent predicted classes from each frame over $W_s$. Aggregating a few frames after adaptation improves stability. We set $W_s=k_c=5$.

The signal combines two sub-signals. Embedding drift $d_{\mathrm{emb}}(i)$ measures how far the current content is from relevant past contexts. Given $\bar{\mathbf{s}}_i$ as the mean of the last $W_s$ semantic scores,
   $$
    d_{\mathrm{emb}}(i) = \frac{1}{|\mathcal{N}_{(\bar{\mathbf{s}}_i, \mathcal{M})}|} \sum_{s^j \in \mathcal{N}_{(\bar{\mathbf{s}}_i, \mathcal{M})}}
    \Big(1 - \text{cos}\big(\bar{\mathbf{s}}_i, \mathbf{s}^{j}\big)\Big),
    \label{eq:embedding_drift}
  $$
where $\mathcal{N}_{(\bar{\mathbf{s}}_i, \mathcal{M})}$ denotes the $J$-nearest memory entries. Using $J$-nearest neighbors rather than the global minimum prevents saturation in long sequences.

Class novelty $d_{\mathrm{nov}}(i)$ detects object categories absent from relevant past contexts. Let $\mathcal{O}_i$ be the union of the top-$k_c$ predicted classes across the current window: 
$$
   d_{\mathrm{nov}}(i) = \frac{\left|\mathcal{O}_i \setminus \{\mathcal{O}^j |(s^j, \mathcal{O}^j) \in \mathcal{N}_{(\bar{\mathbf{s}}_i, \mathcal{M})}\}\right|}
   {|\mathcal{O}_i|}.
$$

The final semantic drift signal combines both: $S_3(i) = \min\!\Big(1,\; \frac{1}{2} \big(d_{\mathrm{emb}}(i) +  d_{\mathrm{nov}}(i)\big)\Big)$. When $S_3>\tau_3$, the semantic content has diverged enough from known contexts, indicating that the adapters are likely inadequate.

\subsection{Hierarchical Fusion}\label{sec:method:fusion}
The three aforementioned signals are combined in a hierarchical logic defined as: 
$$
      \mathcal{H}(S_1, S_2, S_3) =
      \begin{cases}
          (S_2 > \tau_2) \lor (S_3 > \tau_3) & \text{if } S_1 = 1 \\[4pt]
          (S_2 > \tau_2) \land (S_3 > \tau_3) & \text{if } S_1 = 0
      \end{cases}
      \label{eq:hierarchical}
$$
The rationale behind this formulation is that $S_1$ is binary and highly sensitive, as minor variations in the CLIP embeddings immediately activate the signal (e.g., a person approaching the camera). To tackle this, when $S_1$ fires, a confirmation from at least another signal is required. Conversely, when it is silent (no abrupt visual change), both $S_2$ and $S_3$ must agree. Upon triggering, we retrieve the top-$K$ adapters using the mean CLIP embeddings over the last $w$ frames, which is more robust than SemLA's single-frame descriptor.

\section{Experimental Evaluation}
\begin{table*}[ht!]
\scriptsize\addtolength{\tabcolsep}{-1.5pt}
\centering
\caption{Performance comparison in training-free continual test-time adaptation settings
}\label{tab:tf_ctta}
\vspace{-0.2cm}
\label{tab:performance_comparison}
\begin{tabular}{lc|cccccccccccccccc|c}
\multicolumn{2}{c}{}&\multicolumn{16}{c}{ \textbf{t} $\xrightarrow{\hspace{11.4cm}}$}&\\
\toprule
\multicolumn{2}{l|}{\textbf{Test}}& \multicolumn{4}{c|}{Scannet $\rightarrow$ ACDC}& \multicolumn{4}{c|}{Scannet++ $\rightarrow$ ACDC}& \multicolumn{4}{c|}{Scannet $\rightarrow$ MUSES}& \multicolumn{4}{c|}{Scannet++ $\rightarrow$ MUSES} &  \textbf{All}\\

\multicolumn{2}{l|}{\bf{Condition}} & \multicolumn{1}{c}{$\mathbf{E_1^I}$} & \multicolumn{1}{c}{$\mathbf{E_2^I}$} & \multicolumn{1}{c}{$\mathbf{E_3^I}$} & \multicolumn{1}{c|}{$\mathbf{E_4^O}$} & $\mathbf{E_5^I}$ & $\mathbf{E_6^I}$ & $\mathbf{E_7^I}$ & \multicolumn{1}{c|}{$\mathbf{E_8^O}$}& $\mathbf{E_9^I}$ & $\mathbf{E_{10}^I}$ & $\mathbf{E_{11}^I}$ & \multicolumn{1}{c|}{$\mathbf{E_{12}^O}$}& $\mathbf{E_{13}^I}$ & $\mathbf{E_{14}^I}$ & $\mathbf{E_{15}^I}$ & \multicolumn{1}{c|}{$\mathbf{E_{16}^O}$} & \textbf{Mean} \\\midrule

\multirow{2}{*}{Zero-shot} & \textbf{mIoU} &39.0&39.3&39.9&43.1&37.9&36.8&36.3&43.8&38.5&39.0&38.7&35.6&37.5&37.7&36.4&37.6&38.6\\
 & $\mathbf{N_A}$ &0&0&0&0&0&0&0&0&0&0&0&0&0&0&0&0&0\\\midrule

\multirow{2}{*}{Entropy~\cite{emboditta}} & \textbf{mIoU} &44.6&43.4&41.5&42.0&36.4&35.0&35.0&42.3&38.6&38.3&39.9&33.4&36.7&35.7&36.3&35.4&38.4\\
 & $\mathbf{N_A}$ &2&1&1&0&0&0&0&0&0&0&0&0&0&0&0&0&0.3\\\midrule

\multirow{2}{*}{DSS~\cite{simple_signal_domain_shift}} & \textbf{mIoU} &44.8&42.6&40.8&48.5&40.1&37.5&37.1&50.7&42.2&43.2&41.5&41.7&40.0&37.2&38.0&43.5&41.8\\
 & $\mathbf{N_A}$ &1&0&0&1&1&0&0&1&1&0&0&1&1&0&0&1&0.5\\\midrule
\multirow{2}{*}{Periodic} & \textbf{mIoU} &\second{45.2}&\second{44.7}&\best{43.3}&\second{50.6}&\second{42.2}&\second{40.8}&\second{40.6}&\second{51.0}&\second{42.7}&\best{45.6}&\second{44.2}&\second{42.5}&\second{42.8}&\second{41.2}&\best{41.3}&\second{43.5}&\second{43.9}\\
 & $\mathbf{N_A}$ &3&3&3&4&6&6&8&4&2&2&3&2&7&7&7&2&4.3\\\midrule

\multirow{2}{*}{Multi-signal (ours)} & \textbf{mIoU} &\best{45.6}&\best{46.3}&\second{43.0}&\best{52.1}&\best{43.2}&\best{41.2}&\best{41.9}&\best{52.6}&\best{44.0}&\second{45.1}&\best{44.8}&\best{44.6}&\best{43.9}&\best{41.7}&\second{40.8}&\best{45.6}&\best{44.8}\\
 & $\mathbf{N_A}$ &4&3&3&3&7&6&7&2&3&2&3&1&6&6&6&1&3.9\\\midrule\midrule
\multirow{2}{*}{Per-frame~\cite{semla}} & \textbf{mIoU} &46.1&47.6&45.3&53.4&44.0&42.1&42.3&54.1&45.9&47.4&46.0&47.1&44.5&42.4&42.6&48.7&46.2\\
 & $\mathbf{N_A}$ &294&325&350&520&761&773&964&488&302&279&312&185&840&886&807&202&518\\\bottomrule

\multicolumn{19}{l}{}\\[-0.2cm]
\multicolumn{19}{p{0.88\linewidth}}{Results are averaged over 50 sequences of environments, where $\mathbf{E^I_t}$ and $\mathbf{E^O_t}$ refer to indoor and outdoor scenes randomly chosen in the considered datasets. The \best{best} and \second{second} best mIoU performance are evaluated without considering per-frame because of its inefficiency. 
}

\end{tabular}
\vspace{-0.6cm}
\end{table*}
\subsection{Experimental setup and benchmark design}
To evaluate our TF-CTTA approach in open-vocabulary semantic segmentation under domain shift, we design a benchmark that combines a variety of widely used indoor and outdoor datasets. The benchmark covers diverse scene types, label sets, and environmental changes (e.g., weather conditions) to better reflect long-term robotic perception under evolving domain shifts. In addition, we perform evaluation on real-world robotic data.

\noindent
\textbf{Datasets.} Our benchmark is constructed from two indoor and two outdoor datasets.

We use Scannet~\cite{scannet} and Scannet++~\cite{scannetpp}, two large-scale indoor scene understanding datasets with diverse sensing modalities in multiple environment types, such as \texttt{apartment}, \texttt{office}, \texttt{classroom}, \texttt{bedroom}, etc. As in previous works~\cite{continual_adaptation_2d3d,instance_domain_adaptation}, each environment is treated as a separate domain, resulting in 707 and 906 unique domains for Scannet and Scannet++, respectively. For evaluation, we use the NYU40 label space~\cite{nyu40} for Scannet and the top-$100$ classes for Scannet++. 

For outdoor contexts, we use ACDC~\cite{acdc} and MUSES~\cite{muses}, two popular driving datasets where domains are represented by different weather and illumination conditions. More precisely, ACDC defines 5 domains: \texttt{sun}, \texttt{night}, \texttt{rain}, \texttt{fog}, and \texttt{snow}. MUSES extends this to 8 domains: 4 weather conditions (\texttt{clear}, \texttt{rain}, \texttt{fog}, \texttt{snow}) in 2 different light conditions (\texttt{day}, \texttt{night}). Both datasets are evaluated using their 19 standard object categories. 

\noindent\textbf{Implementation details.} We use SED~\cite{sed} as the base model for open-vocabulary semantic segmentation. 
Following~\cite{semla}, we attach a LoRA~\cite{lora} adapter to each linear layer of the model (except the text encoder), training one adapter per domain (1626 total adapters, ${\sim}$10\,MB each). Each adapter is trained until convergence. Since Scannet++ provides data from two hardware configurations (iPhone and DSLR camera), we train adapters using only iPhone data. 

\noindent\textbf{Evaluation protocol and metrics.} To validate our approach, we propose a benchmark in which the model is tested through a sequence of domains, reflecting our continual test-time scenario described in Section~\ref{sec:method:problem_formulation}.

We define a sequence $\mathcal{T}$ alternating 16 domains $\mathbf{E}_t$ randomly chosen from the 4 indoor/outdoor datasets. 
We compose a new label space containing all the object categories needed by the 4 datasets (removing repetitions). This reflects a real-world deployment where the model is prompted with all the objects a robot can find while operating.
Inside the sequence $\mathcal{T}$, we evaluate in a \emph{leave-one-out} fashion: when running in $\mathbf{E}_t$, the corresponding adapter trained in $\mathbf{E}_t$ is discarded, so the model is always tested on unseen scenes. This is a common setup in domain adaptation~\cite{continual_adaptation_2d3d, instance_domain_adaptation} where each environment is considered a different domain. To ensure fairness, the evaluation is repeated over 50 randomly sampled sequences.

Segmentation accuracy is computed using mIoU, while efficiency is measured considering, for each domain, how often the adaptation is triggered (\nadaptations). Minimizing this indicator directly improves inference speed as the adaptation cost is avoided.

\noindent\textbf{Hyper-parameters.} The parameters of the signals introduced in Section~\ref{sec:method:multisignal} are set to $w=21$, $\tau_2=2.0$, $\tau_3=0.15$, and $J=10$, and remain fixed across all datasets. 


\noindent\textbf{Comparative methods.} To evaluate our approach for domain shift detection in TF-CTTA, we compare with the following strategies for triggering adaptation:
\begin{itemize}
    \item Zero-shot: the base model SED~\cite{sed} without adaptation.
    \item Entropy~\cite{emboditta}: adapts when the exponential moving average of the prediction entropy deviates from a post-adaptation baseline beyond a threshold.
    \item DSS~\cite{simple_signal_domain_shift}: triggers adaptation when a domain shift signal (cosine distance between mean feature representations of consecutive batches of images) exceeds a multiple of its moving average. We keep the threshold of the original paper but lower the batch size to 50 frames to make the method more reactive.
    \item Periodic: adaptation is triggered every $\Delta_f = 120$ frames, tuned to match the average number of domain changes (\nadaptations) found by our method.
    \item SemLA~\cite{semla}: the adaptation process is triggered at each frame, with no mechanisms for domain shift detection. 
\end{itemize}

\subsection{Results on training-free continual adaptation}

Table~\ref{tab:performance_comparison} reports the results of our method and the baselines tested in a continual test-time manner. Overall, our method achieves segmentation accuracy (mIoU) close to the SemLA approach that, by performing a per-frame adaptation, represents a performance plateau. At the same time, our approach is substantially more efficient in terms of the number of triggered adaptations (\nadaptations). With only a marginal drop of 1.4 mIoU points compared to SemLA, our method triggers adaptation only 0.8\% of the time.


Compared to all these baselines, our method achieves the best performance in terms of segmentation accuracy in the majority of the splits (13 out of 16), except for $\mathbf{E^I_3}$, $\mathbf{E^I_{10}}$, and $\mathbf{E^I_{15}}$, in which Periodic leads by a margin of 0.3--0.5 mIoU. This demonstrates that our approach is capable of identifying domain shifts in a stream of robotic sensor data for effectively triggering adaptation. 
Notably, our method triggers roughly the same number of adaptations as Periodic but achieves higher segmentation accuracy, as it fires at actual domain transitions instead of fixed intervals. This confirms that when to adapt is as important as how often. It is also important to specify that this is an unrealistic baseline, as the frequency for adaptation has been set to mimic the efficiency reached by our method in this specific benchmark.

The results also demonstrate a substantial improvement of our method over the baselines for domain shift detection, which show poor performance on long-term sequences of robotic perception. Specifically, Entropy~\cite{emboditta} triggers adaptation a few times at the beginning but saturates quickly. This is because entropy is not a reliable signal in a training-free adaptation setting: since there is no online fine-tuning, the model does not progressively gain confidence on the current data, and a single peak in entropy can prevent subsequent adaptation. In contrast, DSS~\cite{simple_signal_domain_shift} identifies clear domain shifts, such as transitions from indoor to outdoor environments and vice versa, but it completely misses smaller domain shifts occurring within the same environment type, which our method captures by combining multiple signals.
\begin{table}[t!]
\centering
\caption{Performance in out-of-distribution settings
}
\vspace{-0.2cm}
\label{tab:results_ood}
\begin{tabular}{lc|cccc|c}
&\multicolumn{1}{c}{}&\multicolumn{4}{c}{ \textbf{t} $\xrightarrow{\hspace{3cm}}$}&\\
\toprule
\textbf{Test} &&\multicolumn{4}{c|}{Scannet++ $\rightarrow$ ACDC}&  \textbf{All}\\

\textbf{Condition} & & \multicolumn{1}{c}{$\mathbf{E_1^I}$} & \multicolumn{1}{c}{$\mathbf{E_2^I}$} & \multicolumn{1}{c}{$\mathbf{E_3^I}$} & \multicolumn{1}{c|}{$\mathbf{E_4^O}$} & \textbf{Mean} \\\midrule
\multirow{2}{*}{Zero-shot} & \textbf{mIoU} &33.6&36.0&34.0&43.0&36.6\\
 & $\mathbf{N_A}$ &0&0&0&0&0\\\midrule
\multirow{2}{*}{Multi-signal (ours)} & \textbf{mIoU} &36.8&38.9&36.2&51.9&41.0\\
 & $\mathbf{N_A}$ &5&3&3&2&3.2\\\midrule\midrule
\multirow{2}{*}{Per-frame~\cite{semla}} & \textbf{mIoU} &36.9&39.4&36.2&53.1&41.4\\
 & $\mathbf{N_A}$ &1084&1066&1202&488&960\\

\bottomrule
\multicolumn{7}{l}{}\\[-0.2cm]
\multicolumn{7}{p{0.85\linewidth}}{Results are averaged over 50 sequences, where $\mathbf{E^I_t}$ and $\mathbf{E^O_t}$ refer to indoor and outdoor scenes.}

\end{tabular}
\vspace{-0.9cm}
\end{table}

\begin{figure*}
    \centering
    \includegraphics[width=0.75\linewidth]
    {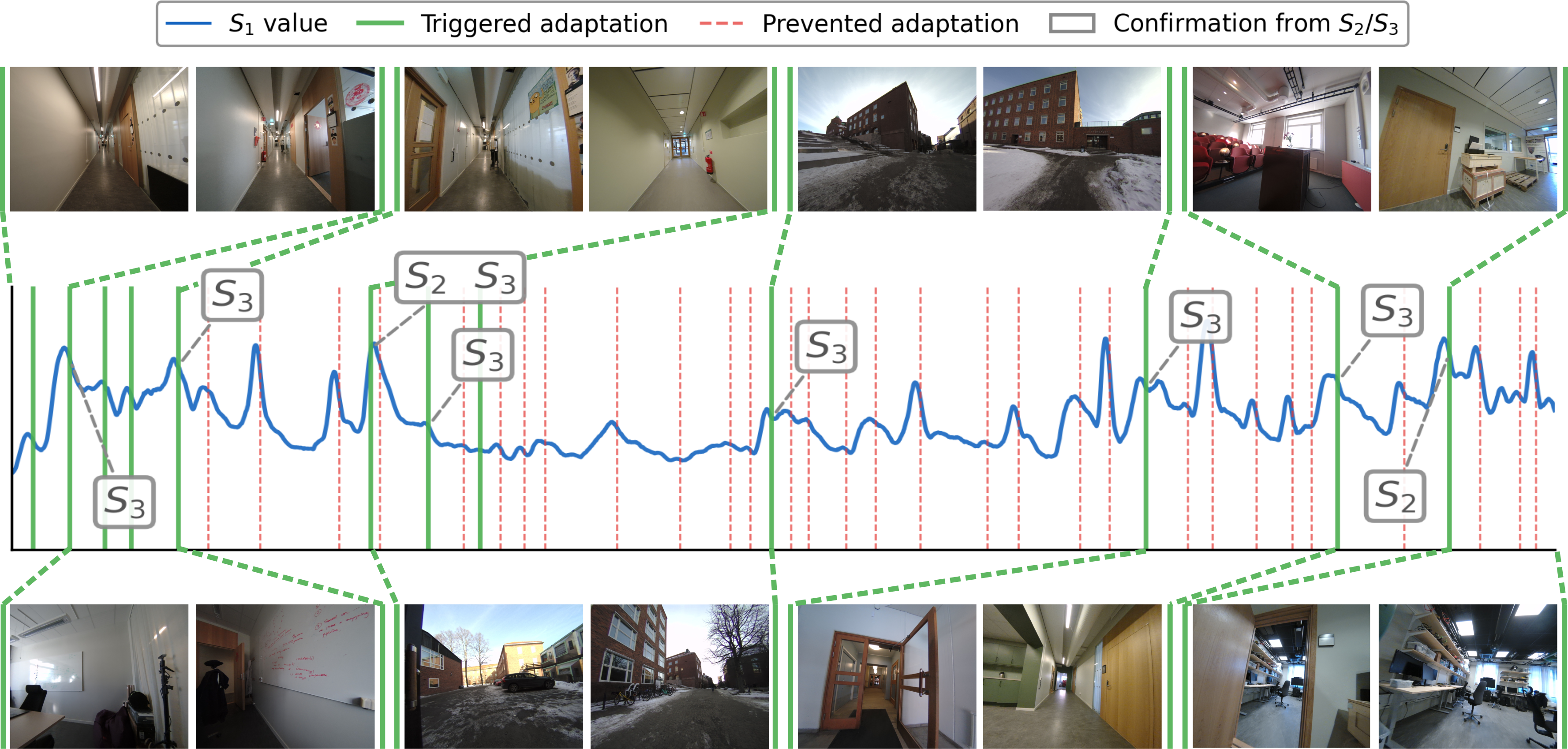}
     \caption{Our multi-signal domain shift detector on a real-world sequence from a robot sensor (Odin) on our university campus, covering indoor and outdoor environments (2735 frames, 12 confirmed adaptations out of 42 $S_1$ activations). Each image pair covers consecutive adaptation events: the first frame is where adaptation is triggered, the second is from the stable period before the next trigger, showing that the domain remains consistent in between.}
    
    \label{fig:robot_sequence_long}
    \vspace{-0.4cm}
\end{figure*}

We perform an additional evaluation in out-of-distribution settings, namely when testing involves a camera model different from those used to train the adapters. This reflects real-world robotic deployment, where the robot's camera often differs from the one used during training. We set up a library of adapters trained on Scannet and Scannet++ with iPhone data, with MUSES as the only outdoor dataset, and test them on Scannet++ using DSLR data (rectified images from a fisheye camera), and on ACDC for outdoor scenes. The results, shown in Table~\ref{tab:results_ood}, first confirm that training-free adaptation remains beneficial in out-of-distribution deployments (+4.4 mIoU over zero-shot). Furthermore, our method achieves performance comparable to per-frame adaptation, with a drop of only 0.4 mIoU, while triggering adaptation for a small fraction of the time (3.2 vs 960).

\subsection{Signal analysis}



\begin{figure}[!t]
    \centering
    \includegraphics[width=\linewidth]{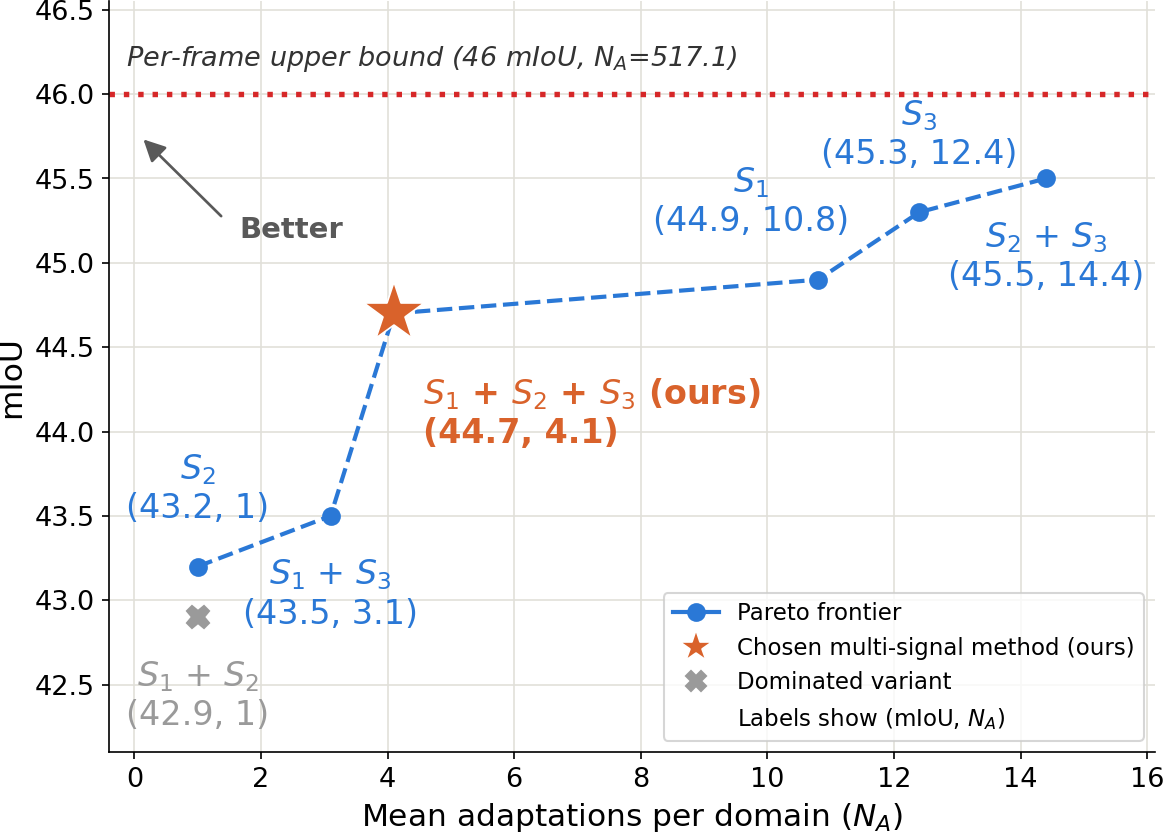}
    \caption{Trade-off between segmentation accuracy and number of adaptations, averaged over 50 sequences, counting 8 domains each. For each point, we report the signals involved and the values of mIoU and $N_A$.}
    \label{fig:pareto}
    \vspace{-0.7cm}
\end{figure}

Fig.~\ref{fig:pareto} shows the accuracy trade-off of individual signals and their combinations. Performance is evaluated as above, averaging over 50 sequences composed of 8 domains instead of 16.
Individually, $S_1$ and $S_3$ detect shifts frequently ($N_A=10.8$ and 12.4, mIoU~44.9 and 45.3), while $S_2$ barely triggers ($N_A=1$, mIoU~43.2), as adapter mismatch builds gradually and needs corroboration from the other signals.
Pairwise combinations expose the role of each signal and how they complement one another. $S_1 + S_2$ is the most conservative combination ($N_A=1$, mIoU~42.9), and is dominated by $S_2$ alone, without $S_3$ the gradual semantic drift is never caught.  $S_1+S_3$ triggers more often ($N_A=3.1$) but obtains only a marginal improvement (mIoU~43.5). $S_2+S_3$ achieves the highest mIoU (45.5) but over-triggers ($N_A=14.4$) since visual change is not considered.
Our full system achieves the best accuracy-to-adaptation trade-off ($N_A=4.1$, mIoU~44.7), staying within 1.3~mIoU of per-frame adaptation (46.0) while triggering 3.5$\times$ fewer adaptations than $S_2+S_3$. This confirms that the signals are individually sensitive but jointly selective, discarding spurious triggers while preserving detection coverage.

\subsection{Real-World Deployment}

We profile the pipeline on an NVIDIA Jetson AGX Orin (32GB, NVMe SSD) mounted on our Spot robot, comparing per-frame adaptation (SemLA) against our triggered adaptation. Both are tested at two power levels (60W and 30W) and two model sizes to reflect different deployment scenarios. The image size is set to 224$\times$224 and 384$\times$384 for CLIP and SED, respectively. 
Table~\ref{tab:efficiency} reports the per-component latency: domain navigation computed using CLIP (Domain nav.), the inference with the segmentation model (Seg. model), and adaptation cost (Adapt. cost). The domain shift signals introduce no additional model inference, operating on embeddings and predictions already available in the pipeline.
The results clearly show that adaptation is the bottleneck. While inference time can be reduced using smaller backbones, the adaptation cost is harder to cut as it is bounded by weight loading and merging. Even with our optimized adapter loading procedure compared to the original implementation~\cite{semla}, which directly swaps adapters without creating new layers for storing them, each adaptation still takes on average 550ms at full power and 700ms at low power, more than CLIP and SED inference combined (except for the configuration with large models and low power). As a result, per-frame adaptation cuts throughput by 1.6--3.3$\times$, i.e., from 2.6 to 1.1 FPS at full power and from 0.8 to 0.5 FPS at low power with the large model. Since this cost is largely architecture-independent, minimizing the number of adaptations is a necessary condition for real-time deployment: with our triggered adaptation and the small model, the pipeline reaches 4.2 FPS at full power (3.2$\times$ faster compared to the per-frame).

\begin{table}[t!]
\scriptsize
\centering
\caption{Latency of the per-frame vs. ours adaptations (\nadaptations $= 4$)}

\vspace{-0.2cm}
\label{tab:efficiency}

\begin{tabular}{cc|ccccc}
\toprule
\makecell{Power\\mode} & \makecell{Models\\dim.} &\makecell{Domain\\nav.} & \makecell{Seg.\\model} & \makecell{Adapt.\\cost} & \makecell{FPS\\(per-frame)}&\makecell{FPS\\(\nadaptations $= 4$)}\\\midrule
\multirow{2}{*}{High} & Large& 150ms& 230ms&550ms& 1.1&\best{2.6}\\
& Small & 90ms & 140ms & 550ms& 1.3 & \best{4.2}\\\midrule
\multirow{2}{*}{Low}& Large & 350ms & 840ms & 700ms & 0.5& \best{0.8}\\
& Small & 110ms&490ms&700ms&0.7&\best{1.7}\\\bottomrule

\multicolumn{7}{l}{}\\[-0.2cm]
\multicolumn{7}{p{0.92\linewidth}}{Performance evaluated on a Jetson AGX Orin using two power modalities: 60W (High) and 30W (Low).}
\end{tabular}
\vspace{-0.7cm}
\end{table}


We deploy our method on a real-world sequence captured using the Odin sensor~\cite{odin} while navigating indoor/outdoor in our university campus. The adapter library is trained on the benchmark datasets, and all signal parameters remain identical to the benchmark experiments, except $\tau_2=1.5$ (lowered from 2.0), as the Odin image embeddings lie further from all adapters, raising $\delta_{\mathrm{base}}$ so that the adapter mismatch signal never triggers at the default value. The camera captures at 10 FPS, but as the pipeline throughput on Jetson is 4.2 FPS (Table~\ref{tab:efficiency}), we subsample the stream by 2, processing 2735 frames to simulate real-time operation.

Fig.~\ref{fig:robot_sequence_long} shows $S_1$ and the adaptation events over the full sequence. $S_1$ alone triggers 42 times (red dashed), but the hierarchical fusion confirms only 12 (green), rejecting $71\%$ of $S_1$ activations as false positives during stable periods. $S_3$ (semantic drift) confirms most triggers, while $S_2$ (adapter mismatch) confirms them when the active adapters become inadequate for the current scene. The confirmed adaptations correspond to visually distinct domain transitions: corridor changes, indoor-to-outdoor passages, and room type shifts (office, cinema, and laboratory). On the Jetson hardware, 12 adaptations add only 7s of overhead compared to 25 minutes under per-frame adaptation.  This experiment validates that the detector generalizes to unseen environments and a different sensor with a single parameter change, complementing the quantitative out-of-distribution evaluation in Table~\ref{tab:results_ood}, whose test images are of the same type as the Odin ones (rectified from a fisheye camera).

\section{Conclusions}
In this paper, we propose a pipeline for training-free domain adaptation with a multi-signal domain shift detector to trigger adaptation only when needed,  substantially improving efficiency (3.3$\times$ FPS) while maintaining accuracy.

In future work, we plan to investigate new strategies for indexing and merging adapters to improve segmentation accuracy and further reduce adaptation cost.


\bibliographystyle{IEEEtran}
\bibliography{references}

@InProceedings{semla,
    author    = {Qorbani, Reza and Villani, Gianluca and Panagiotakopoulos, Theodoros and Colomer, Marc Botet and H\"arenstam-Nielsen, Linus and Segu, Mattia and Dovesi, Pier Luigi and Karlgren, Jussi and Cremers, Daniel and Tombari, Federico and Poggi, Matteo},
    title     = {Semantic Library Adaptation: LoRA Retrieval and Fusion for Open-Vocabulary Semantic Segmentation},
    booktitle = {Proc. of CVPR},
    month     = {June},
    year      = {2025},
    pages     = {9804-9815}
}

@online{odin,
author = "Manifold Tech",
title = "Manifold Odin 1",
url = "https://www.manifoldtech.cn/products/Odin1",
addendum = ""
}

@INPROCEEDINGS{multimodal_ctta,
  author={Liu, Yan and Zhu, Hongyuan and Zhang, Ye and Lei, Yinjie and Guo, Yulan},
  booktitle={Proc. of ICRA}, 
  title={Multi-Modality Test-Time Adaptation for Semantic Segmentation in Robotic Perception}, 
  year={2025},
  volume={},
  number={},
  pages={16963-16969},
  doi={10.1109/ICRA55743.2025.11127565}}

@inproceedings{nyu40,
  title={Indoor segmentation and support inference from rgbd images},
  author={Silberman, Nathan and Hoiem, Derek and Kohli, Pushmeet and Fergus, Rob},
  booktitle={Proc. of ECCV},
  pages={746--760},
  year={2012},
  organization={Springer}
}

@InProceedings{ctta_dynamic_selection,
    author    = {Wang, Yanshuo and Hong, Jie and Cheraghian, Ali and Rahman, Shafin and Ahmedt-Aristizabal, David and Petersson, Lars and Harandi, Mehrtash},
    title     = {Continual Test-Time Domain Adaptation via Dynamic Sample Selection},
    booktitle = {Proc. of WACV},
    year      = {2024},

}

@InProceedings{finetuning_vfm,
    author    = {Wortsman, Mitchell and Ilharco, Gabriel and Kim, Jong Wook and Li, Mike and Kornblith, Simon and Roelofs, Rebecca and Lopes, Raphael Gontijo and Hajishirzi, Hannaneh and Farhadi, Ali and Namkoong, Hongseok and Schmidt, Ludwig},
    title     = {Robust Fine-Tuning of Zero-Shot Models},
    booktitle = {Proc. of CVPR},
    month     = {June},
    year      = {2022},
    pages     = {7959-7971}
}

@ARTICLE{survey_domain_adaotation,
  author={Schwonberg, Manuel and Niemeijer, Joshua and Termöhlen, Jan-Aike and schäfer, Jörg P. and Schmidt, Nico M. and Gottschalk, Hanno and Fingscheidt, Tim},
  journal={IEEE Access}, 
  title={Survey on Unsupervised Domain Adaptation for Semantic Segmentation for Visual Perception in Automated Driving}, 
  year={2023},
  volume={11},
  number={},
  pages={54296-54336},
  }

@ARTICLE{perception_for_authonomy,
  author={Kunze, Lars and Hawes, Nick and Duckett, Tom and Hanheide, Marc and Krajník, Tomáš},
  journal={IEEE ROBOT AUTOM LETT}, 
  title={Artificial Intelligence for Long-Term Robot Autonomy: A Survey}, 
  year={2018},
  volume={3},
  number={4},
  pages={4023-4030},
  doi={10.1109/LRA.2018.2860628}}

@inproceedings{uniform_adapter_merging,
  title={Training-free model merging for multi-target domain adaptation},
  author={Li, Wenyi and Gao, Huan-ang and Gao, Mingju and Tian, Beiwen and Zhi, Rong and Zhao, Hao},
  booktitle={Proc. of ECCV},
  pages={419--438},
  year={2024},
  organization={Springer}
}

@InProceedings{sed,
    author    = {Xie, Bin and Cao, Jiale and Xie, Jin and Khan, Fahad Shahbaz and Pang, Yanwei},
    title     = {SED: A Simple Encoder-Decoder for Open-Vocabulary Semantic Segmentation},
    booktitle = {Proc. of CVPR},
    month     = {June},
    year      = {2024},
    pages     = {3426-3436}
}

@InProceedings{clip,
  title = 	 {Learning Transferable Visual Models From Natural Language Supervision},
  author =       {Radford, Alec and Kim, Jong Wook and Hallacy, Chris and Ramesh, Aditya and Goh, Gabriel and Agarwal, Sandhini and Sastry, Girish and Askell, Amanda and Mishkin, Pamela and Clark, Jack and Krueger, Gretchen and Sutskever, Ilya},
  booktitle = 	 {Proc. of ICML},
  pages = 	 {8748--8763},
  year = 	 {2021},
  volume = 	 {139}
}

@inproceedings{lora,
  title={Lora: Low-rank adaptation of large language models.},
  author={Hu, Edward J and Shen, Yelong and Wallis, Phillip and Allen-Zhu, Zeyuan and Li, Yuanzhi and Wang, Shean and Wang, Liang and Chen, Weizhu and others},
  booktitle={Proc. of ICLR},
  volume={1},
  number={2},
  pages={3},
  year={2022}
}

@ARTICLE{dual_ctta,
  author={Tian, Yuntong and Li, Kang and He, Tianyang and Wan, Liang and Heng, Pheng-Ann and Feng, Wei},
  journal={IEEE Transactions on Image Processing}, 
  title={Dual Domain-Attribute Learning Framework With Asynchronous Adapters for Continual Test-Time Adaptation}, 
  year={2026},
  volume={35},
  number={},
  pages={376-387},
}

@inproceedings{onda,
  title={Online domain adaptation for semantic segmentation in ever-changing conditions},
  author={Panagiotakopoulos, Theodoros and Dovesi, Pier Luigi and H{\"a}renstam-Nielsen, Linus and Poggi, Matteo},
  booktitle={Proc. of ECCV},
  year={2022},
  organization={Springer}
}

@InProceedings{real_time_ctta,
    author    = {Colomer, Marc Botet and Dovesi, Pier Luigi and Panagiotakopoulos, Theodoros and Carvalho, Joao Frederico and H\"arenstam-Nielsen, Linus and Azizpour, Hossein and Kjellstr\"om, Hedvig and Cremers, Daniel and Poggi, Matteo},
    title     = {To Adapt or Not to Adapt? Real-Time Adaptation for Semantic Segmentation},
    booktitle = {Proc. of ICCV},
    month     = {October},
    year      = {2023},
    pages     = {16548-16559}
}

@InProceedings{hybrid_tta,
    author    = {Park, Hyewon and Park, Hyejin and Ko, Jueun and Min, Dongbo},
    title     = {Hybrid-TTA: Continual Test-time Adaptation via Dynamic Domain Shift Detection},
    booktitle = {Proc. of ICCV},
    month     = {October},
    year      = {2025},
    pages     = {2877-2886}
}

@inproceedings{muses,
  title={Muses: The multi-sensor semantic perception dataset for driving under uncertainty},
  author={Br{\"o}dermann, Tim and Bruggemann, David and Sakaridis, Christos and Ta, Kevin and Liagouris, Odysseas and Corkill, Jason and Van Gool, Luc},
  booktitle={Proc. of ECCV},
  pages={21--38},
  year={2024},
  organization={Springer}
}

@InProceedings{acdc,
    author    = {Sakaridis, Christos and Dai, Dengxin and Van Gool, Luc},
    title     = {ACDC: The Adverse Conditions Dataset With Correspondences for Semantic Driving Scene Understanding},
    booktitle = {Proc. of ICCV},
    month     = {October},
    year      = {2021},
    pages     = {10765-10775}
}

@InProceedings{scannet,
author = {Dai, Angela and Chang, Angel X. and Savva, Manolis and Halber, Maciej and Funkhouser, Thomas and Niessner, Matthias},
title = {ScanNet: Richly-Annotated 3D Reconstructions of Indoor Scenes},
booktitle = {Proc. of CVPR},
month = {July},
year = {2017}
}

@InProceedings{scannetpp,
    author    = {Yeshwanth, Chandan and Liu, Yueh-Cheng and Nie{\ss}ner, Matthias and Dai, Angela},
    title     = {ScanNet++: A High-Fidelity Dataset of 3D Indoor Scenes},
    booktitle = {Proc. of ICCV},
    month     = {October},
    year      = {2023},
    pages     = {12-22}
}

@inproceedings{ecotta,
  title={Ecotta: Memory-efficient continual test-time adaptation via self-distilled regularization},
  author={Song, Junha and Lee, Jungsoo and Kweon, In So and Choi, Sungha},
  booktitle={Proc. of CVPR},
  pages={11920--11929},
  year={2023}
}

@inproceedings{ccotta,
  title={Controllable continual test-time adaptation},
  author={Shi, Ziqi and Lyu, Fan and Liu, Ye and Shang, Fanhua and Hu, Fuyuan and Feng, Wei and Zhang, Zhang and Wang, Liang},
  booktitle={Proc. of ICME},
  pages={1--6},
  year={2025},
  organization={IEEE}
}

@INPROCEEDINGS{cotta,
  author={Wang, Qin and Fink, Olga and Van Gool, Luc and Dai, Dengxin},
  booktitle={Proc. of CVPR}, 
  title={Continual Test-Time Domain Adaptation}, 
  year={2022},
  volume={},
  number={},
  pages={7191-7201},
}

@inproceedings{vida,
 author = {Liu, Jiaming and Yang, Senqiao and Jia, Peidong and Zhang, Renrui and Lu, Ming and Guo, Yandong and Xue, Wei and Zhang, Shanghang},
 booktitle = {Proc. of ICLR},
 editor = {B. Kim and Y. Yue and S. Chaudhuri and K. Fragkiadaki and M. Khan and Y. Sun},
 pages = {48396--48417},
 title = {ViDA: Homeostatic Visual Domain Adapter for Continual Test Time Adaptation},
 year = {2024}
}

@INPROCEEDINGS{simple_signal_domain_shift,
  author={Chakrabarty, Goirik and Sreenivas, Manogna and Biswas, Soma},
  booktitle={Proc. of ICCVW}, 
  title={A Simple Signal for Domain Shift}, 
  year={2023},
  volume={},
  number={},
  pages={3569-3576}}

@misc{emboditta,
  title={EmbodiTTA: Resource-Efficient Test-Time Adaptation for Embodied Visual Systems},
  author={Ma, Xiao and Kwon, Young D and Ma, Dong},
  year={2025}
}

@ARTICLE{continual_adaptation_2d3d,
  author={Frey, Jonas and Blum, Hermann and Milano, Francesco and Siegwart, Roland and Cadena, Cesar},
  journal={IEEE ROBOT AUTOM LETT}, 
  title={Continual Adaptation of Semantic Segmentation Using Complementary 2D-3D Data Representations}, 
  year={2022}
}

@inproceedings{instance_domain_adaptation,
      title={Instance-Guided Unsupervised Domain Adaptation for Robotic Semantic Segmentation}, 
      author={Michele Antonazzi and Lorenzo Signorelli and Matteo Luperto and Nicola Basilico},
      year={2026},
    booktitle={Proc. of ICRA}
}
\end{document}